\documentclass[letterpaper, 10 pt, conference]{ieeeconf}  

\IEEEoverridecommandlockouts                              

\title{\LARGE \bf
ExtraGS: Enhancing Endoscopic View Extrapolation via Diffusion-Guided 3D Gaussian Splatting}

\author{Cheng-Tai Hsieh$^{1}$, Jiwei Shan$^{2}$, Han Fang$^{3}$, Jianshu Hu$^{3}$, Tao Ni$^{4}$, \\ Lijun Han$^{1}$, Yutong Ban$^{3}$, Shing Shin Cheng$^{2}$, and Hesheng Wang$^{1,\dagger}$%
\thanks{$\dagger$ This work was supported by the Fundamental and Interdisciplinary Disciplines Breakthrough Plan of the Ministry of Education of China (JYB2025XDXM117), the National Natural Science Foundation of China (62225309, U24A20278, 62361166632, 62403311, and 62503322), and the AI for Science Seed Program of Shanghai Jiao Tong University (2025AI4SQY06). (Corresponding author: Hesheng Wang.)}%
\thanks{$^{1}$Cheng-Tai Hsieh, Lijun Han, and Hesheng Wang are with the School of Automation and Intelligent Sensing, Shanghai Jiao Tong University, and the Shanghai Key Laboratory of Navigation and Location-Based Services, Shanghai, China. (E-mail: wanghesheng@sjtu.edu.cn)}%
\thanks{$^{2}$Jiwei Shan and Shing Shin Cheng are with the Department of Mechanical and Automation Engineering and the T Stone Robotics Institute, The Chinese University of Hong Kong, Hong Kong.}%
\thanks{$^{3}$Han Fang, Jianshu Hu, and Yutong Ban are with the Global College, Shanghai Jiao Tong University, Shanghai, China.}%
\thanks{$^{4}$Tao Ni is with Shanghai Ninth People's Hospital, Shanghai Jiao Tong University School of Medicine, Shanghai, China.}%
}

\usepackage{amssymb}
\usepackage{graphicx}
\usepackage{cite}
\usepackage{multirow}
\usepackage{booktabs}
\usepackage{amsmath}
\usepackage{graphicx}
\usepackage{subfigure}
\usepackage{textcomp}
\usepackage{color}
\usepackage{tabularx}
\usepackage{multirow}
\usepackage{xcolor}
\usepackage{bm}
\definecolor{indigo}{rgb}{0.29, 0.0, 0.51}
\definecolor{gray}{rgb}{0.5, 0.5, 0.5} 
\usepackage[colorlinks, linkcolor=blue, anchorcolor=blue, citecolor=blue]{hyperref}
\begin{document}

\maketitle
\thispagestyle{empty}
\pagestyle{empty}

\begin{abstract}
Robot-assisted minimally invasive surgery (MIS) critically depends on reliable endoscopic perception for navigation and safety. However, conventional endoscopes provide only a limited field of view, leaving large portions of surrounding anatomy unobserved. Recent neural rendering approaches, such as Neural Radiance Fields and 3D Gaussian Splatting, enable novel view synthesis from endoscopic videos, but their reliance on sparse observations often leads to severe artifacts when extrapolating beyond the training trajectory.
In this work, we propose ExtraGS, a framework for enhancing endoscopic view extrapolation via diffusion-guided 3D Gaussian Splatting. Starting from an initial reconstruction, we introduce an uncertainty-guided virtual camera sampling strategy to actively explore blind spots and maximize information gain. The rendered views from these sampled locations are refined using a diffusion model to recover plausible anatomical structures, producing pseudo observations that guide further optimization. To prevent the generated content from degrading reliable regions, we adopt a confidence-weighted fine-tuning strategy when incorporating these pseudo observations.
Extensive experiments on multiple public endoscopic datasets demonstrate that ExtraGS significantly reduces extrapolation artifacts and achieves state-of-the-art performance in endoscopic novel view synthesis. Code is available at https://github.com/IRMVLab/ExtraGS.
\end{abstract}

\section{INTRODUCTION}
Endoscopy is the gold standard for screening and diagnosing diseases, such as polyps and early-stage cancers, in luminal organs like the gastrointestinal tract \cite{lieberman2012guidelines}. However, traditional endoscopes have a restricted field of view (FOV) due to hardware constraints and the narrow anatomical space. This limitation poses significant clinical challenges. For example, during a procedure, physicians can only observe the local tissue directly in front of the camera. This restricted view often creates visual blind spots, potentially leading to missed diagnoses \cite{menon2014commonly}. Additionally, it forces surgeons to rely heavily on their spatial memory to mentally reconstruct the 3D anatomy of the examined region. Therefore, generating novel views from endoscopic videos and providing perspectives beyond the physical camera's limits may have potential value for preoperative planning, intraoperative navigation, and retrospective diagnosis.

\begin{figure}[t]
    \centering
    \includegraphics[width=0.95\columnwidth]{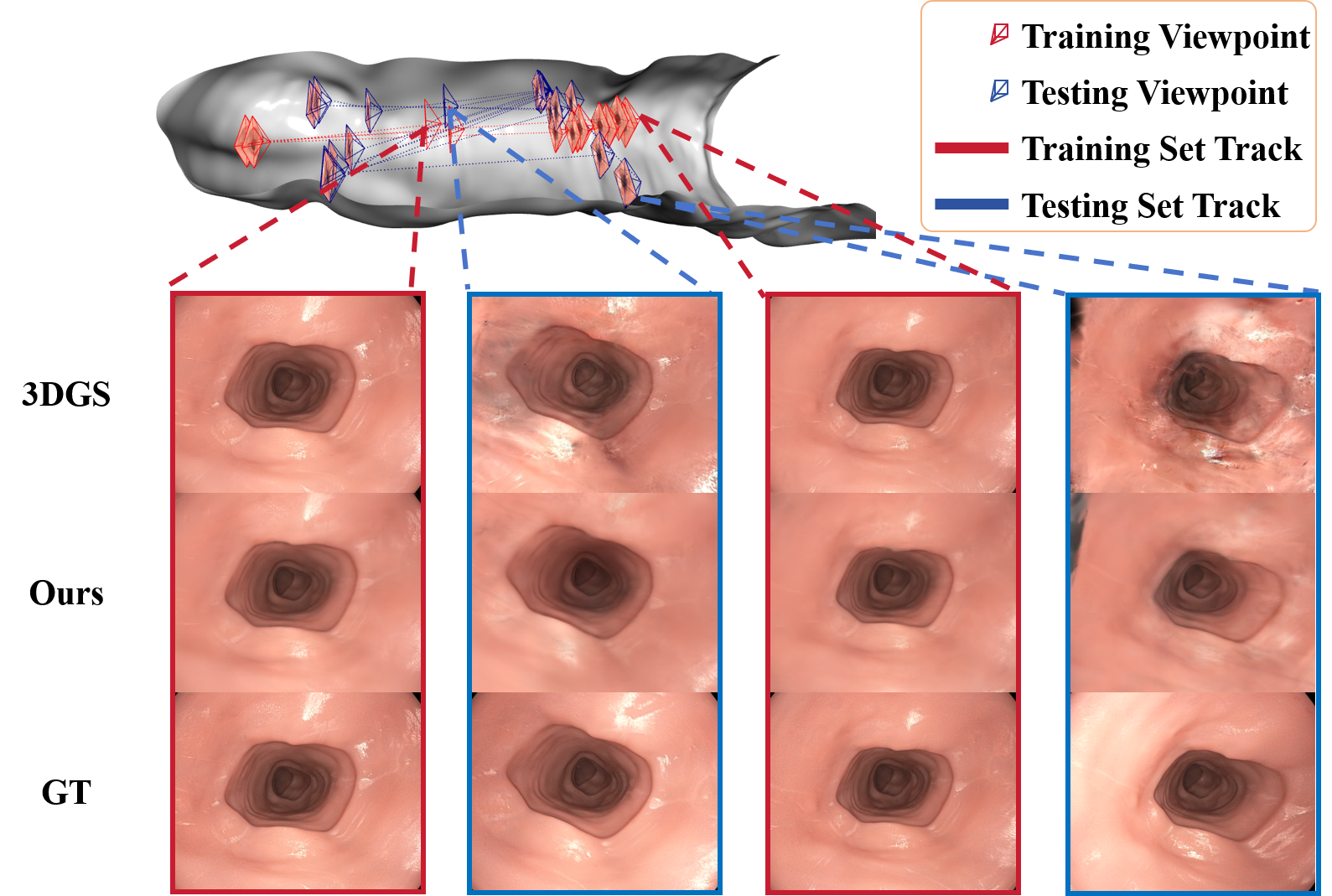}
    \caption{\textbf{Extra-GS enables exploration-driven reconstruction.} Top: Cross-sectional view of the dataset mesh illustrating training (red) and testing (blue) camera trajectories. Bottom: Representative renderings from off-trajectory viewpoints. While standard 3DGS suffers from severe artifacts when deviating from the training path, our method maintains structural completeness and high fidelity under large viewpoint shifts.}
    \label{fig:fig1}
    \vspace{-5mm}
\end{figure}

Recently, differentiable rendering techniques, including Neural Radiance Fields (NeRF) \cite{mildenhall2020nerf} and 3D Gaussian Splatting (3DGS) \cite{kerbl20233dgs}, have achieved success in novel view synthesis. These techniques have been adapted to endoscopic videos for 3D/4D reconstruction \cite{wang2022neural,zha2023endosurf,bonilla2024gaussian,shan2025deformable,shan2025uw,batlle2023lightneus,shan20264d}, SLAM \cite{shan2024enerf,shan2024dds,wu2025endoflow,shan2026nrgs}, and novel view synthesis \cite{guo2024uc}. To compensate for insufficient multi-view constraints under the narrow FOV, existing methods introduce geometric priors such as depth \cite{wang2022neural}, normals \cite{bonilla2024gaussian}, or 2D pixel trajectories \cite{wu2025endoflow,shan20264d}. These methods perform well in view interpolation, rendering views with minor offsets from the input viewpoints. However, as shown in Fig.~\ref{fig:fig1}, their performance degrades significantly during view extrapolation, where the virtual camera moves far from these reference views. This failure stems from the training phase: the limited physical viewpoint coverage leaves many anatomical regions unobserved. Without actual image inputs, neither basic photometric losses nor auxiliary geometric priors can provide effective supervision to optimize the 3D representation (e.g., 3D Gaussian attributes) in these blind spots. Consequently, when the virtual camera moves far from the reference views to expose these unoptimized regions, the rendered images exhibit severe geometric distortions and chaotic artifacts, rendering them clinically unreliable.


To address the aforementioned challenges, we propose \textbf{ExtraGS}, a novel framework for {E}nhancing Endoscopic View \textbf{Extra}polation via Diffusion-Guided 3D \textbf{G}aussian \textbf{S}platting. Our core insight is that to overcome the lack of supervision in unobserved areas, a generative prior is essential to synthesize plausible content. Leveraging the strong inpainting capabilities and rich anatomical knowledge captured by diffusion model \cite{rombach2022high}, we integrate a 2D diffusion prior into the endoscopic 3DGS optimization process to effectively resolve view extrapolation failures.
Specifically, our technical pipeline begins with training a base 3DGS model on the sparse input views. We then introduce an uncertainty estimation mechanism to quantify and locate under-observed regions. Guided by this uncertainty, a novel virtual camera sampling strategy actively targets these blind spots to maximize information gain. Next, we render initial, artifact-prone images from these sampled viewpoints and process them through the diffusion model to restore high-fidelity anatomical structures. Finally, these enhanced images serve as pseudo-ground truth to update the 3DGS model. To prevent the generated content from degrading accurately reconstructed areas, we employ a confidence-weighted fine-tuning strategy. Extensive evaluations on multiple open-source endoscopic datasets demonstrate that ExtraGS significantly reduces artifacts and holes during view extrapolation. Our contributions are summarized as follows:
\begin{itemize}
    \item We propose {ExtraGS}, a novel framework integrating 2D diffusion priors into endoscopic 3DGS to solve severe artifact issues during large-baseline view extrapolation.
    \item We introduce an uncertainty-guided virtual camera sampling mechanism to actively target blind spots and maximize information gain.
    \item We design a confidence-weighted fine-tuning strategy to incorporate diffusion-generated content, filling unobserved areas while preserving accurate 3D structures.
    \item Evaluation on multiple open-source endoscopic datasets demonstrates state-of-the-art performance on both novel view synthesis and large-baseline view extrapolation tasks, supported by comprehensive ablations.
\end{itemize}

\section{Related Works}

\subsection{Differentiable Rendering in Endoscopy}

Differentiable rendering techniques, notably Neural Radiance Fields (NeRF) \cite{mildenhall2020nerf} and 3D Gaussian Splatting (3DGS) \cite{kerbl20233dgs}, revolutionize 3D reconstruction and novel view synthesis in endoscopy. 
Initially, researchers widely adapt NeRF for deformable soft tissue reconstruction \cite{wang2022neural,zha2023endosurf,shan2025uw}, high-fidelity static 3D reconstruction \cite{batlle2023lightneus} novel view synthesis (NVS), \cite{guo2024uc} and dense visual SLAM. Despite impressive synthesis quality, NeRF's heavy reliance on dense MLP evaluations via volumetric ray marching leads to prohibitively slow training and rendering speeds, which hinders real-time intraoperative applications.
To overcome these computational bottlenecks, 3D Gaussian Splatting (3DGS) emerges as a powerful alternative.
\par\newpage\noindent
By representing scenes with explicit 3D Gaussians and employing an efficient tile-based rasterizer, 3DGS bypasses expensive ray-marching. Consequently, researchers rapidly adopt 3DGS in endoscopy for similar tasks spanning from deformable soft tissue modeling \cite{huang2024endo,shan2025deformable,shan20264d} to real-time endoscopy SLAM \cite{shan2024enerf,shan2024dds,wu2025endoflow,shan2026nrgs} , which achieves state-of-the-art rendering quality at real-time frame rates. Furthermore, to compensate for the narrow endoscopic field of view inherent in these tasks, current methods often incorporate geometric priors, such as depth or normals.

Despite these promising advancements, existing methods fundamentally fail during large-baseline view extrapolation, producing severe geometric distortions and artifacts. To address this limitation, we propose ExtraGS, a novel framework that integrates 2D diffusion priors into endoscopic 3DGS to ensure robust and artifact-free rendering, even during large-baseline view extrapolation.
      
\subsection{Diffusion Priors for Novel View Synthesis}

Since traditional optimization-based methods cannot hallucinate unobserved regions, recent works increasingly leverage 2D diffusion priors for novel view synthesis (NVS). Methods like DreamFusion \cite{pooledreamfusion} and ReconFusion \cite{wu2024reconfusion} employ Score Distillation Sampling (SDS) or fine-tuned diffusion models to recover 3D representations from sparse views. 
More closely related to our task, several methods leverage diffusion priors for large-baseline view extrapolation. For instance, ExtraNeRF \cite{shih2024extranerf} and ExploreGS \cite{kim2025exploregs} utilize image inpainting and video diffusion priors to synthesize unseen areas. However, these methods are primarily tailored for natural scenes and overlook the unique characteristics of endoscopic environments. Directly applying them to surgical videos often yields suboptimal results.

\section{METHOD}
\subsection{Overview}
\label{sec:method_overview}
\begin{figure*}[t]
    \centering
    \includegraphics[width=0.95\textwidth]{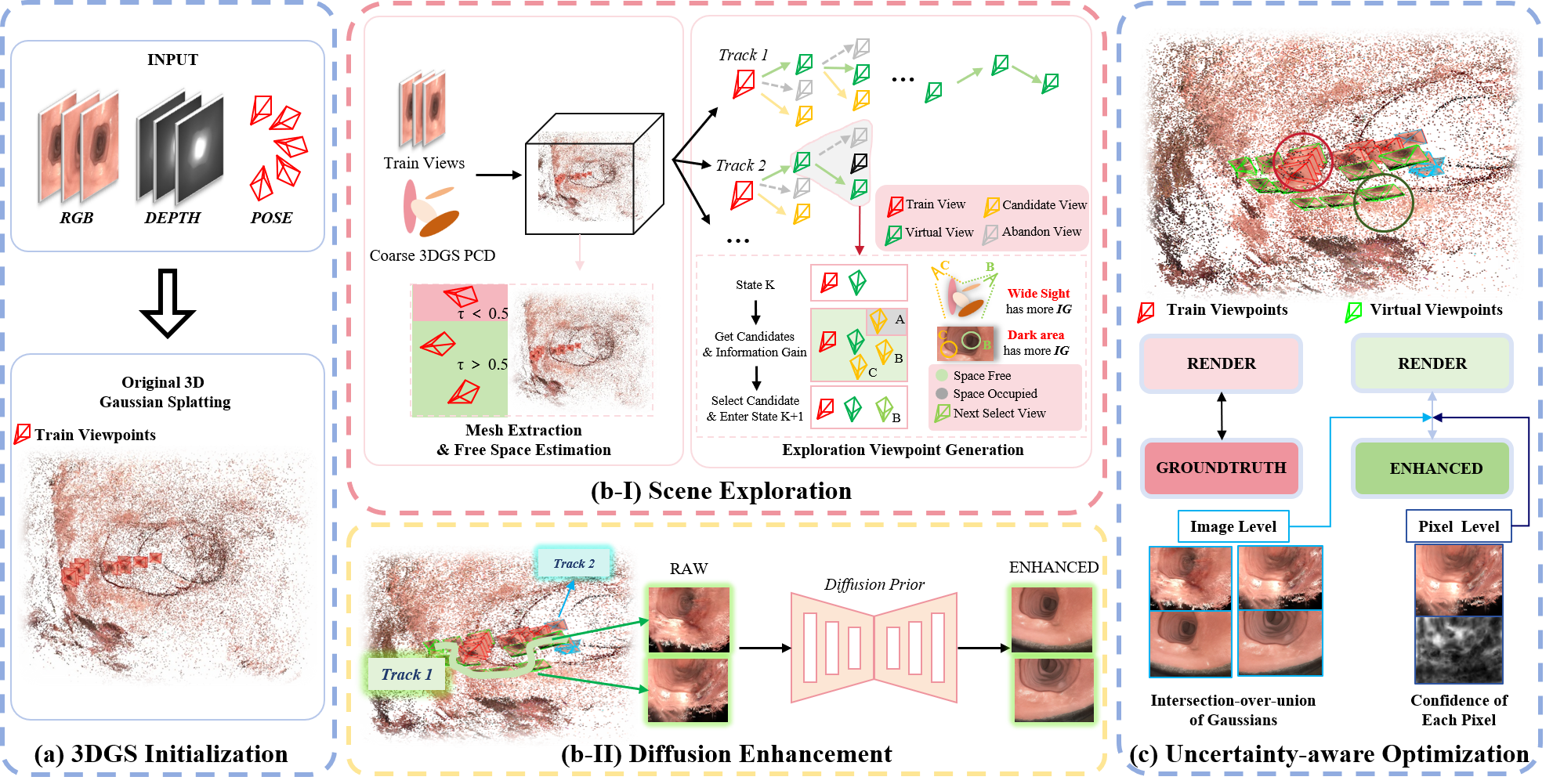}
    \caption{\textbf{Overview of the Extra-GS framework.} Our proposed pipeline consists of three primary stages: (a) \textbf{Scene Initialization:} A baseline 3D Gaussian Splatting (3DGS) representation is constructed utilizing available endoscopic RGB and depth images, alongside camera poses estimated via COLMAP. (b) \textbf{Pseudo-Observation Generation:} Novel viewpoints are synthesized through strategic virtual camera sampling and subsequently enhanced using diffusion-prior refinement. (c) \textbf{Uncertainty-Aware Fine-Tuning:} The 3D Gaussian scene is jointly optimized using both the real and generated pseudo-observations. Uncertainty information is incorporated during this optimization to effectively guide the learning process, ultimately improving the structural completeness and rendering robustness of the model.}
    \label{fig:overview}
\end{figure*}

Given a set of posed endoscopic RGB images and their corresponding depth maps, we aim to achieve robust large-baseline view extrapolation in surgical scenes. Our method builds upon 3D Gaussian Splatting (3DGS) and leverages diffusion priors. Therefore, we first review these key techniques in Sec.~\ref{sec:method_prelim}. The overall pipeline of ExtraGS is illustrated in Fig.~\ref{fig:overview}. The process consists of four main steps. First, we optimize a coarse 3D Gaussian representation $\mathcal{G}_{\text{c}}$ using the input data. From this representation, we extract a surface mesh to determine the feasible region for virtual camera exploration (Sec.~\ref{sec:method_stage1_init}). Then, we sample virtual camera trajectories within this feasible region using a strategy tailored to surgical scenes, enabling more comprehensive coverage of the entire scene (Sec.~\ref{sec:method_sample}). Along these trajectories, we sample virtual viewpoints, render corresponding images using the coarse  $\mathcal{G}_{\text{c}}$, and refine them using a video diffusion model to produce high-quality pseudo-observations (Sec.~\ref{sec:method_diffusion}). Finally, we fine-tune the coarse 3D Gaussians using both the original inputs and the pseudo-observations via a confidence-weighted optimization strategy (Sec.~\ref{sec:method_stage3}).

\newpage
\subsection{Preliminaries}
\label{sec:method_prelim}

\subsubsection{{3D Gaussian Splatting}}
\label{sec:method_prelim_3dgs}
3D Gaussian Splatting (3DGS) \cite{kerbl20233dgs} represents a scene explicitly as a set of anisotropic 3D Gaussians, denoted as \(\mathcal{G}=\{g_j\}_{j=1}^{N_G}\). Each Gaussian \(g_j\) is parameterized by its center \(\bm{\mu}_j \in \mathbb{R}^3\), opacity \(\alpha_j\), view-dependent color coefficients \(\mathbf{c}_j\) (typically represented with spherical harmonics), and a covariance matrix \(\bm{\Sigma}_j\). The covariance is commonly decomposed into a scaling matrix \(\mathbf{S}_j\) and a rotation matrix \(\mathbf{R}_j\). The spatial distribution of a Gaussian is defined as
\begin{equation}
g_j(\mathbf{x})=\exp\!\left(-\frac{1}{2}(\mathbf{x}-\bm{\mu}_j)^{T}\bm{\Sigma}_j^{-1}(\mathbf{x}-\bm{\mu}_j)\right).
\end{equation}

To render an image, each 3D Gaussian is projected onto the image plane via splatting~\cite{ewa}. Under the camera transformation, the projected mean and covariance are computed as \(\bm{\mu}_j^{\prime}=\mathbf{J}\mathbf{W}\bm{\mu}_j\) and \(\bm{\Sigma}_j^{\prime}=\mathbf{J}\mathbf{W}\bm{\Sigma}_j\mathbf{W}^{T}\mathbf{J}^{T}\), where \(\mathbf{W}\) denotes the world-to-camera transformation and \(\mathbf{J}\) represents the Jacobian of the local affine approximation of the perspective projection.
The rendered color at pixel \(\mathbf{p}\) is obtained by front-to-back alpha compositing over the depth-ordered set \(\mathcal{N}(\mathbf{p})\) of projected Gaussians:
\begin{equation}
\hat{\mathbf{I}}(\mathbf{p})=
\sum_{j\in\mathcal{N}(\mathbf{p})}
\mathbf{c}_j\,\alpha_j(\mathbf{p})
\prod_{k=1}^{j-1}\bigl(1-\alpha_k(\mathbf{p})\bigr),
\end{equation}
where \(\alpha_j(\mathbf{p})\) denotes the opacity contribution of the \(j\)-th Gaussian at pixel \(\mathbf{p}\).
During training, the Gaussian parameters are optimized by minimizing the discrepancy between the rendered image \(\hat{I}\) and the ground-truth image \(I\):
\begin{equation}
\label{eq:loss_train_gs}
\mathcal{L}_{\text{pho}} 
=
(1-\lambda)\,\|\hat{I}-I\|_1
+\lambda\,\mathcal{L}_{\text{D-SSIM}},
\end{equation}
where \(\|\cdot\|_1\) denotes the \(L_1\) loss, $\text{SSIM}$ is the Structural Similarity Index Measure, and \(\lambda\) is a balancing weight.

\subsubsection{Diffusion Prior}
\label{sec:method_prelim_diffusion}
Renderings from virtual viewpoints typically preserve reliable content in well-observed regions,
but exhibit severe degradations (e.g., holes, floaters, and inconsistent textures) under large viewpoint changes.
A \emph{conditional video diffusion prior} can be used to transform a degraded rendering $I^{V}$
into a realistic pseudo-observation $I^{P}$ while preserving trustworthy cues.

Let $z_0=\mathcal{E}(I^{P})$ denote the clean latent code of the target frame (or a short video clip),
where $\mathcal{E}(\cdot)$ is a VAE encoder.
Given a diffusion timestep $t\in\{1,\dots,T\}$, the forward process gradually perturbs $z_0$
with Gaussian noise according to a variance schedule $\{\alpha_t\}_{t=1}^{T}$, where $\alpha_t\in(0,1)$
and $\bar{\alpha}_t=\prod_{s=1}^{t}\alpha_s$:
\begin{equation}
q(z_t \mid z_0)=\mathcal{N}\!\left(\sqrt{\bar{\alpha}_t}\,z_0,\; (1-\bar{\alpha}_t)\mathbf{I}\right),
\end{equation}
where $z_t$ is the noisy latent at step $t$ and $\mathbf{I}$ is the identity matrix.
The reverse process then predicts the injected noise and iteratively removes it using a conditional denoiser
$\epsilon_{\psi}$ parameterized by $\psi$:
\begin{equation}
z_{t-1}=
\frac{1}{\sqrt{\alpha_t}}
\left(
z_t-\frac{1-\alpha_t}{\sqrt{1-\bar{\alpha}_t}}\,
\epsilon_{\psi}(z_t,t,\mathcal{C})
\right)
+\sigma_t \mathbf{n},
\quad \mathbf{n}\sim\mathcal{N}(0,\mathbf{I}),
\end{equation}
where $\sigma_t$ controls the sampling variance and $\mathcal{C}$ denotes the conditioning inputs.

Concretely, the conditional diffusion model $\Phi_{\psi}$ performs iterative denoising in latent space
and decodes the result into an enhanced pseudo-observation:
\begin{equation}
I^{P} = \Phi_{\psi}\!\left(I^{V},\, I^{\mathrm{ref}},\, \mathbf{condition}\right),
\end{equation}
where $I^{\mathrm{ref}}$ denotes the nearest real observation serving as the visual anchor,
and $\mathbf{condition}$ denotes optional conditioning context.
For video diffusion, the same formulation applies to spatiotemporal latent tensors, enabling temporally coherent enhancement across frames.

\subsection{Scene Initialization and Feasible Region Estimation}
\label{sec:method_stage1_init}

We first optimize a coarse 3D Gaussian representation $\mathcal{G}_{\text{c}}$ using the training set, following the 3DGS optimization by minimizing Eq.~\eqref{eq:loss_train_gs}. Based on this initial reconstruction, we estimate a feasible spatial region of the surgical scene to constrain subsequent virtual viewpoint exploration.

Specifically, we extract a surface mesh from $\mathcal{G}_{\text{c}}$ and combine it with the input RGB images to compute a bounding volume that approximates the reconstructable workspace. This boundary restricts viewpoint exploration to regions supported by the input evidence and prevents sampling from drifting into areas beyond the observed scene. Within this boundary, we further identify occupied anatomy and retain only free-space locations for virtual camera placement via rasterization-based occupancy estimation. We construct an occupancy grid $\mathbf{O}\in\mathbb{R}^{S\times S\times S}$, where $S$ denotes the grid resolution along each axis, and estimate the visibility of each 3D Gaussian from the training views using transmittance values computed by the Gaussian rasterizer. As in ExploreGS \cite{kim2025exploregs}, the top-3 visibility scores are averaged to estimate visibility at each location. A voxel is classified as occupied if its visibility is below a threshold $\tau$ (set to $0.5$ in our experiments), and otherwise regarded as free space. The feasible region $\Omega$ is defined as the free-space subset inside the bounding volume, and all subsequent virtual viewpoint sampling is restricted to $\Omega$.


\begingroup
\setlength{\abovedisplayskip}{2.5pt plus 1pt minus 0.5pt}
\setlength{\belowdisplayskip}{2.5pt plus 1pt minus 0.5pt}
\subsection{Virtual Exploration View Sampling}
\label{sec:method_sample}

To improve scene coverage and complement regions not observed in the input views, we actively sample virtual camera trajectories within the feasible region $\Omega$. Building on the search strategy of ExploreGS \cite{kim2025exploregs}, we progressively construct virtual trajectories while incorporating illumination-derived uncertainty, direction-aware coverage, and pose-smoothness control tailored to endoscopic scenes.

\subsubsection{Search initialization and candidate generation}
Let $\mathcal{T}_n=\{v_{n,1},\dots,v_{n,L}\}$ denote the $n$-th virtual trajectory of length $L$, where $v_{n,t}$ represents the camera state at step $t$, and the current state corresponds to the most recently appended viewpoint. We initialize $N_{\mathrm{Tr}}$ trajectories from input viewpoints selected at spatially uniform intervals. At each step, the current state is expanded into a set of candidate views defined by camera motion primitives in $SE(3)$:
\begin{equation}
\mathcal{C}_t=\{v_t^{(a)}\}_{a=1}^{A}, \qquad A=14,
\end{equation}
including six translational moves, four pure camera rotations, and four orbiting rotations around the look-at point. A candidate $v_t^{(a)}$ is discarded if it (i) lies outside the feasible region $\Omega$, (ii) falls into an occupied voxel of the occupancy grid $\mathbf{O}$, or (iii) is closer than a safety margin to the reconstructed Gaussians.
These constraints prevent the trajectory from drifting away from the tubular anatomy or producing invalid pseudo-observations from anatomically occupied regions.

\subsubsection{Coverage gain with illumination-derived uncertainty}
To evaluate candidate views, we measure information gain based on view coverage. We maintain a binary coverage map
\begin{equation}
\mathbf{M}\in\{0,1\}^{N_G\times D},
\end{equation}
where $N_G$ denotes the number of Gaussians and $D$ is the number of discretized viewing directions. Each entry $\mathbf{M}[j,k]=1$ indicates that Gaussian $g_j$ has been observed from the $k$-th direction bin, and the map is initialized using the input viewpoints. For a candidate view $v_t^{(a)}$, let $\mathcal{J}(v_t^{(a)})$ denote the set of visible Gaussians and $k(v_t^{(a)})$ its viewing-direction index. The map update is
\begin{equation}
\mathbf{M}_t[j,k]=1,\qquad \forall j\in\mathcal{J}(v_t^{(a)}),\; k=k(v_t^{(a)}),
\end{equation}
and the newly acquired coverage is
\begin{equation}
\Delta \mathbf{M}_t=\mathbf{M}_t-\mathbf{M}_{t-1}.
\end{equation}
The direction-aware coverage gain is the number of newly covered Gaussian--direction pairs:
\begin{equation}
\label{eq:ig_coverage}
\mathrm{IG}_{c}(v_t^{(a)})=
\sum_{j=1}^{N_G}\sum_{k=1}^{D}\Delta \mathbf{M}_t[j,k],
\end{equation}
where $\mathrm{IG}_{c}$ measures the number of newly observed Gaussian--direction pairs. This coverage term is subsequently modulated by illumination-derived uncertainty for candidate ranking.

\subsubsection{Illumination-aware exploration reweighting}
Endoscopic images often contain severely dark regions caused by distance-dependent attenuation and directional lighting. 
Such regions are typically less constrained by the input views and thus more likely to be insufficiently reconstructed. 
To encourage exploration toward these areas, we modulate the coverage gain with an illumination-aware weight derived from the darkest region in the current image. Let $\mathbf{d}^{\mathrm{dark}}_t\in\mathbb{S}^2$ denote the world-space unit ray obtained by back-projecting the darkest image location (or the center of the darkest local patch) of the current image through the camera at $v_t$. 
For each candidate $v_t^{(a)}$ with camera center $\mathbf{c}_t^{(a)}$, the exploration direction relative to the current state $\mathbf{c}_t$ is defined as
\begin{equation}
\mathbf{d}^{(a)}_t=
\frac{\mathbf{c}_t^{(a)}-\mathbf{c}_t}
{\|\mathbf{c}_t^{(a)}-\mathbf{c}_t\|_2}.
\end{equation}
The illumination-aware weight is computed using their cosine similarity:
\begin{equation}
\label{eq:ill_weight}
w_{\mathrm{ill}}(v_t^{(a)})=
1+\lambda_l \,\max\!\left(0,\,
\langle \mathbf{d}^{\mathrm{dark}}_t,\mathbf{d}^{(a)}_t\rangle
\right).
\end{equation}
This weighting favors motions aligned with the darkest direction, encouraging exploration toward darker and potentially under-reconstructed regions. The illumination-aware information gain is therefore
\begin{equation}
\mathrm{IG}(v_t^{(a)})=
w_{\mathrm{ill}}(v_t^{(a)})\cdot \mathrm{IG}_{c}(v_t^{(a)}).
\end{equation}

\subsubsection{Fold-aware pose variation control}
Endoscopic scenes often contain complex folds and concave structures, where abrupt viewpoint changes may lead to unstable rendering. 
To encourage smooth camera motion during trajectory expansion, we measure the inter-step pose variation as
\begin{equation}
\Delta R_t^{(a)}=\angle\!\left(R_t^{-1}R_t^{(a)}\right),
\end{equation}
where $R_t$ and $R_t^{(a)}$ denote the rotations of the current and candidate views. 
Candidates with $\Delta R_t^{(a)}>\tau_R$ are treated as warning states and penalized by
\begin{equation}
\mathrm{IG}(v_t^{(a)})\leftarrow \eta \,\mathrm{IG}(v_t^{(a)}),\qquad 0<\eta<1.
\end{equation}
This discourages abrupt viewpoint changes and stabilizes trajectory expansion.

\subsubsection{{Trajectory construction and termination}}
At each step, all valid candidates in $\mathcal{C}_t$ are evaluated and ranked according to $\mathrm{IG}(\cdot)$. 
The candidate with the highest score is appended to the current trajectory as the next state, while the remaining candidates are stored in a priority queue for future branch expansion. 
The global coverage map $\mathbf{M}$ is shared across all trajectories to avoid redundant exploration across different paths. The search terminates in either of the following cases: 
(1) no valid candidate can be expanded before the trajectory reaches length $L$, in which case the current branch is pruned and the search resumes from the next-best candidate in the queue; or 
(2) the current trajectory reaches the predefined length $L$. 
This process repeats until $N_{\mathrm{Tr}}$ virtual trajectories are constructed. 
The resulting viewpoints are then used to render images for the subsequent diffusion-based refinement stage.

\subsection{View Enhancing with Diffusion Prior}
\label{sec:method_diffusion}

Given the sampled virtual trajectories, we convert the rendered virtual views into supervision for reconstruction refinement. 
For each virtual viewpoint $v \in \mathcal{T}_n$, we first render a coarse image
\begin{equation}
I^{V}(v)=\mathcal{R}\!\left(\mathcal{G}_{\text{c}}, v\right),
\end{equation}
where $\mathcal{R}(\cdot)$ denotes Gaussian rendering from the initialized scene $\mathcal{G}_{\text{c}}$.
Since these viewpoints may lie beyond the support of the input views, the rendered images $I^{V}(v)$ may contain blur, floaters, missing structures, or illumination artifacts.
To address this issue, we employ the diffusion prior as a \emph{virtual-view enhancement module}. 
Instead of interpolating between neighboring input frames, the enhancement is conditioned on the input view closest to the queried virtual camera to guide view completion under large viewpoint changes. 
The pseudo-observation at $v$ is generated as
\begin{equation}
I^{P}(v)=
\Phi_{\psi}\!\left(
I^{V}(v),\,
I^{T}\!\left(v_{\mathrm{nn}}(v)\right),\,
\mathbf{c}(v)
\right),
\end{equation}
where $v_{\mathrm{nn}}(v)$ denotes the nearest input viewpoint to $v$, and $\mathbf{c}(v)$ represents auxiliary conditioning associated with the virtual query, such as relative pose encoding. 
This formulation preserves reliable structures in the rendered view while correcting degraded or unsupported regions using the nearest input view as reference.
Applying this procedure to all sampled viewpoints produces the pseudo-observation set
\begin{equation}
\mathcal{V}_{P}=
\left\{
I^{P}(v),\,v
\right\},
\end{equation}
which is used for the subsequent joint fine-tuning stage.

\subsection{Joint Fine-tuning}
\label{sec:method_stage3}

We fine-tune the coarse Gaussian scene representation $\mathcal{G}_0$
using both the input views $\mathcal{V}_T=\{(I^T, V^T)\}$
and the pseudo-observations $\mathcal{V}_P=\{(I^P, V^V)\}$.
A key challenge is that diffusion-enhanced pseudo images may introduce subtle inconsistencies,
which can degrade already well-reconstructed regions.
Inspired by confidence-weighted refinement \cite{liu20243dgs-enhancer}, we mitigate this issue at two levels: an image-level weight measures complementary view coverage, whereas a pixel-level weight localizes content corrected by diffusion. Together, they emphasize new virtual-view information while real observations anchor well-reconstructed regions.

\subsubsection{Image-level confidence via Gaussian IoU (G-IOU)}
Let $V^V$ be a virtual viewpoint and $V^{\text{ref}}$ its nearest input viewpoint.
IoU, short for intersection over union, generally quantifies the overlap between two sets. We adopt the G-IOU definition of ExploreGS~\cite{kim2025exploregs} to measure the normalized overlap of the Gaussians visible from these two viewpoints.
Using the rasterizer, we extract the visible Gaussian index set $\mathcal{G}_{\text{vis}}(\cdot)$
and define the co-visible set
\begin{equation}
\label{eq:gcovis}
\mathcal{G}_{\text{covis}}(V^V, V^{\text{ref}})
=
\mathcal{G}_{\text{vis}}(V^V)\cap \mathcal{G}_{\text{vis}}(V^{\text{ref}}).
\end{equation}
The Gaussian IoU is defined as
\begin{equation}
\label{eq:giou}
\mathrm{GIOU}(V^V, V^{\text{ref}})
=
\frac{\left|\mathcal{G}_{\text{covis}}(V^V, V^{\text{ref}})\right|}
{\left|\mathcal{G}_{\text{vis}}(V^{\text{ref}})\right|},
\end{equation}
where normalization by the reference-view set gives the fraction of its visible Gaussians that remain visible from the virtual viewpoint. The image-level confidence is then computed as
\begin{equation}
\label{eq:uimg}
U_{\text{img}}(V^V)=1-\mathrm{GIOU}(V^V, V^{\text{ref}}).
\end{equation}
Here, $U_{\text{img}}\in[0,1]$ is an image-wise scalar; lower overlap yields a larger weight because the virtual view provides more complementary coverage.

\subsubsection{Pixel-level confidence via perceptual difference}
We compute a spatial LPIPS feature-distance map between the coarse rendering $I^V$ and its diffusion-enhanced pseudo-observation $I^P$, and upsample it to the image resolution:
\begin{equation}
\label{eq:upixel}
U_{\text{pixel}}(x,y)=\mathcal{U}\!\left(\mathrm{LPIPS}(I^V, I^P)\right)(x,y),
\end{equation}
where $\mathcal{U}(\cdot)$ denotes upsampling. Thus, $U_{\text{pixel}}\in\mathbb{R}^{H\times W}$ emphasizes pixels changed most by diffusion, focusing refinement on degraded or missing content.

\subsubsection{Training-view loss}
For training viewpoints, we use the standard 3DGS reconstruction objective:
\begin{equation}
\label{eq:l_train}
\mathcal{L}_{\mathcal{V}_T}
=
\|I^T-\hat{I}(V^T;\mathcal{G})\|_1
+
\mathcal{L}_{\text{D-SSIM}}\!\left(I^T,\hat{I}(V^T;\mathcal{G})\right).
\end{equation}
Here, $\hat I(V;\mathcal G)=\mathcal R(\mathcal G,V)$ is the current 3DGS rendering; this loss anchors content supported by real observations.

\subsubsection{Virtual-view loss with two-level confidence}
For pseudo observations, we apply confidence-weighted supervision:
\begin{equation}
\label{eq:l_virtual}
\mathcal{L}_{\mathcal{V}_P}
=
U_{\text{img}}\cdot
\Big(
\|I^P-\hat{I}\|_1\odot U_{\text{pixel}}
+
\mathcal{L}_{\text{D-SSIM}}\!\left(I^P,\hat{I}\right)
\Big),
\end{equation}
where $\odot$ is pixel-wise multiplication, $U_{\text{img}}$ is broadcast over the image, $I^P$ is the target, and $\hat I=\hat I(V^V;\mathcal G)$ is the current rendering. Jointly optimizing the real- and virtual-view objectives transfers diffusion-restored information into $\mathcal G$ without discarding observed content.
\endgroup

\begin{figure*}[t]
    \centering
    \setlength{\tabcolsep}{1pt} 
    \renewcommand{\arraystretch}{0} 
    \begin{tabular}{ccccc}
        \textbf{GT} & \textbf{3DGS} & \textbf{UCNeRF} & \textbf{GaussianPancakes}  & \textbf{Ours} \\[2pt]

        \includegraphics[width=0.158\linewidth]{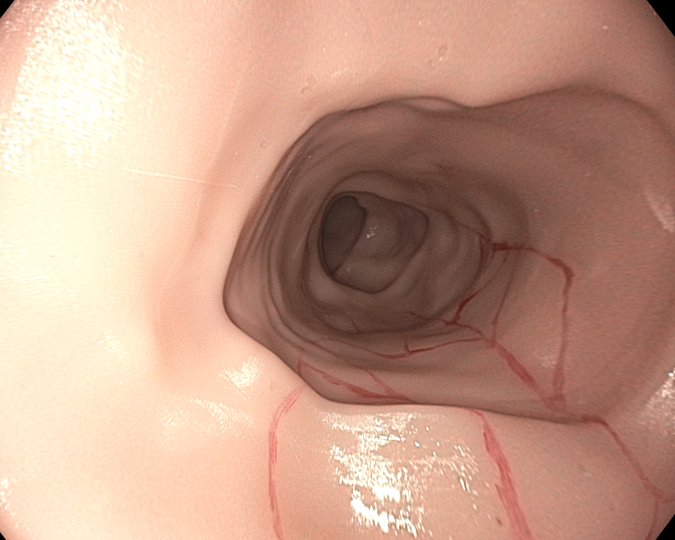} &
        \includegraphics[width=0.158\linewidth]{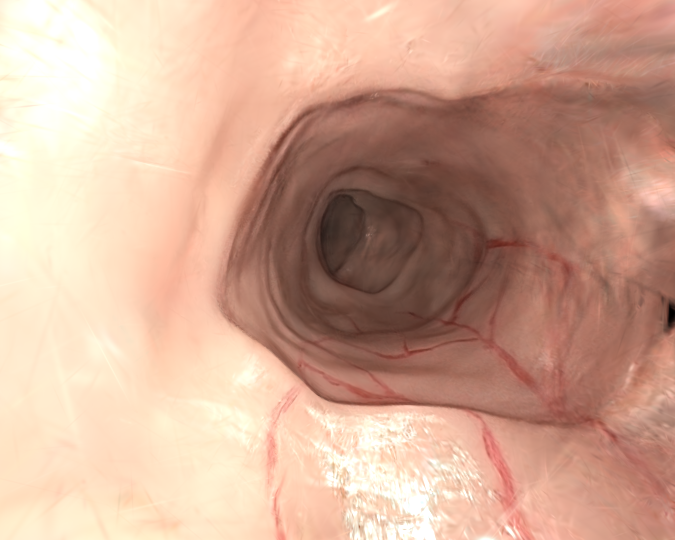} &
        \includegraphics[width=0.158\linewidth]{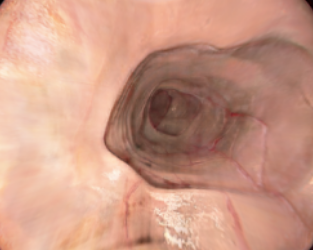} &
        \includegraphics[width=0.158\linewidth]{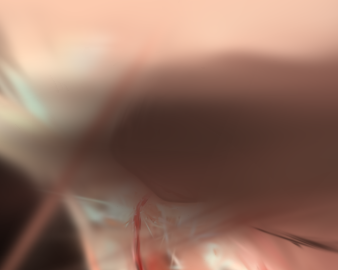}  &
        \includegraphics[width=0.158\linewidth]{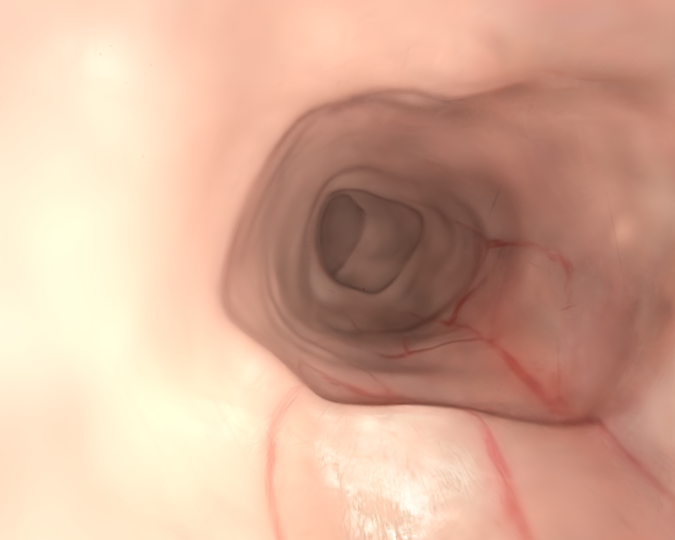} \\[1pt]

        \includegraphics[width=0.158\linewidth]{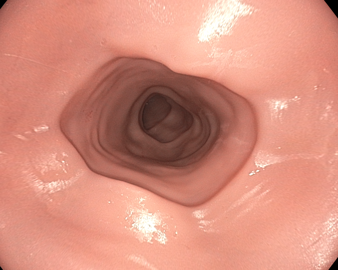} &
        \includegraphics[width=0.158\linewidth]{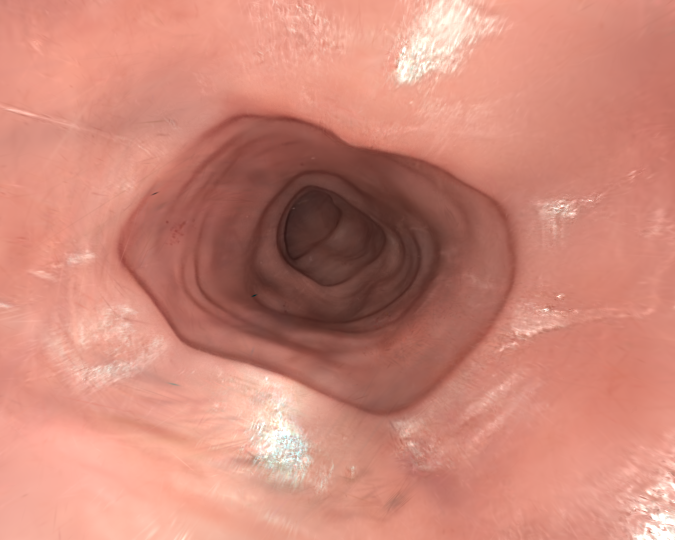} &
        \includegraphics[width=0.158\linewidth]{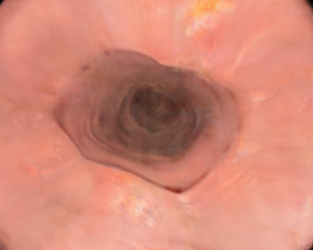} &
        \includegraphics[width=0.158\linewidth]{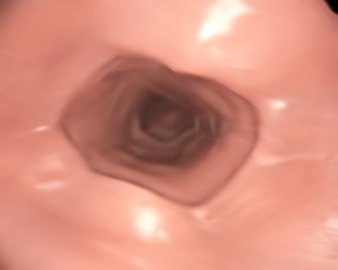}  &
        \includegraphics[width=0.158\linewidth]{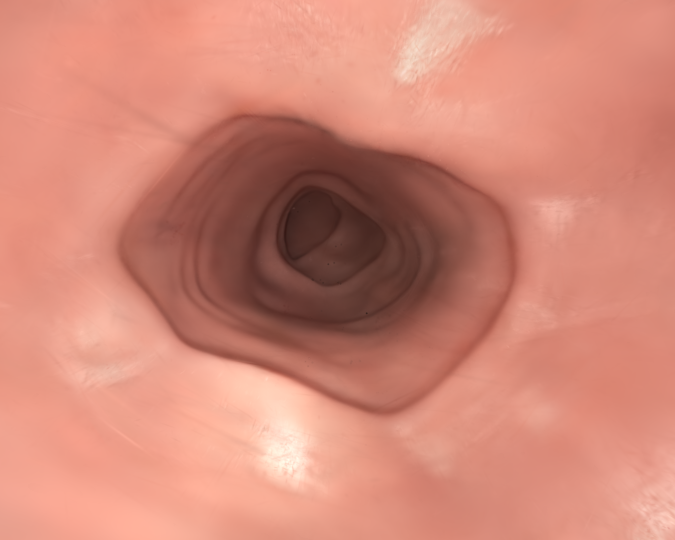} \\[1pt]

        \includegraphics[width=0.158\linewidth]{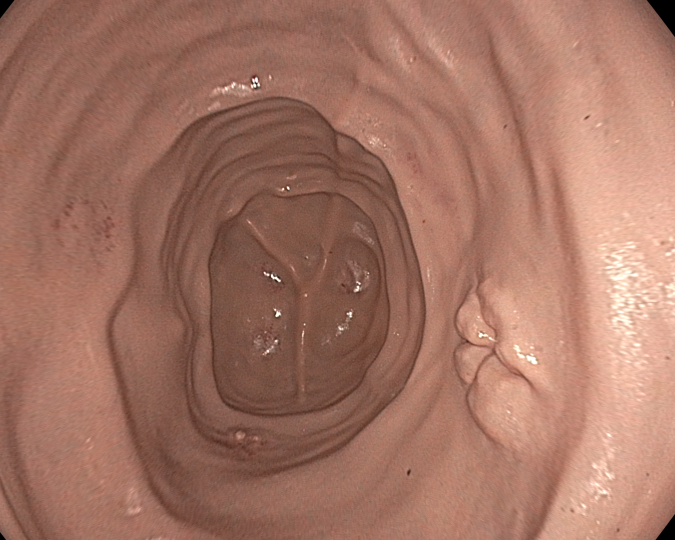} &
        \includegraphics[width=0.158\linewidth]{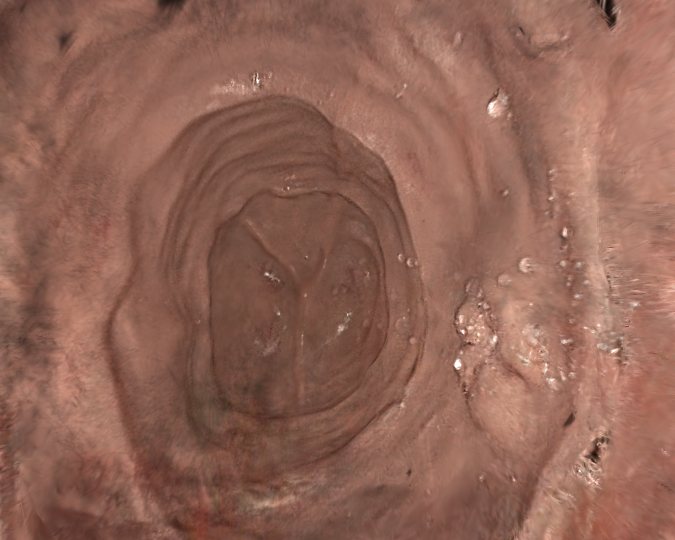} &
        \includegraphics[width=0.158\linewidth]{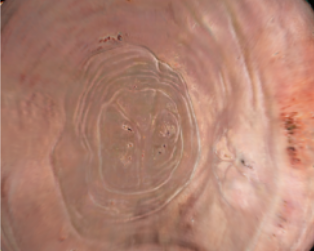} &
        \includegraphics[width=0.158\linewidth]{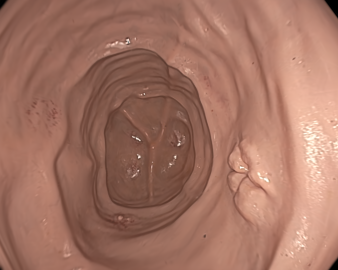}  &
        \includegraphics[width=0.158\linewidth]{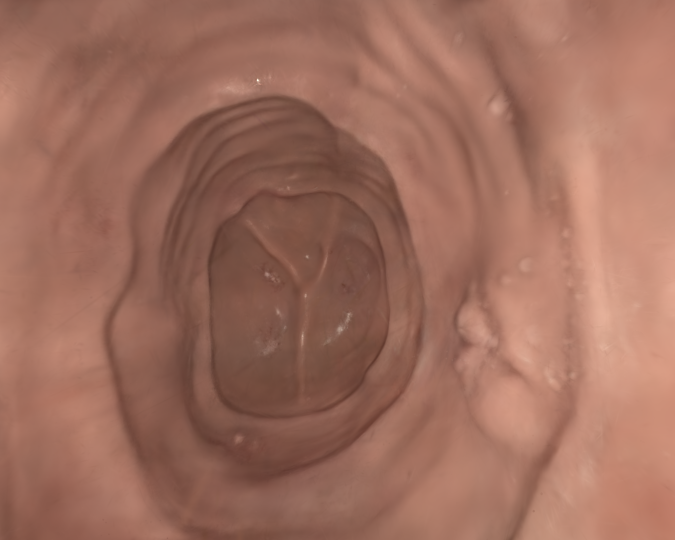} \\[1pt]
    \end{tabular}
    \caption{\textbf{Qualitative comparison of novel view synthesis on the C3VD dataset.} Rendered novel views from baseline methods (3DGS, UCNeRF, GaussianPancakes) and our proposed ExtraGS are shown alongside the Ground Truth. The visualizations illustrate the rendering performance of each approach in reconstructing challenging endoscopic environments characterized by complex tissue topologies, fine vascular networks, and specular highlights.}
    \label{fig:qualitative_nerfbusters}
\end{figure*}

\newpage
\section{Experiments}
\subsection{Experimental Setup}
\label{sec:exp_setup}
\subsubsection{Datasets} We evaluate our method on {C3VDv2} dataset~\cite{golhar2025c3vdv2}. 
C3VDv2 contains colonoscopic sequences with calibrated poses and depth information, characterized by narrow tubular structures, repetitive folds, and severe self-occlusions. 

\subsubsection{Baselines and Metrics}
We evaluate our method against a series of representative reconstruction approaches for endoscopic scenes, including 3DGS~\cite{kerbl20233dgs}, UC-NeRF~\cite{guo2024uc}, and GaussianPancakes~\cite{bonilla2024gaussian}. 
In all experiments, we follow the original protocols of the compared methods as closely as possible. 
This includes adapting the dataset organization and input data format when required by different implementations, while preserving the same underlying scenes, observations, and evaluation settings to ensure fair comparison. For quantitative comparison, we use the peak signal-to-noise ratio (PSNR), structural similarity index (SSIM) and learning-based perceptual image patch similarity (LPIPS) as metrics to evaluate reconstruction performance.
\subsubsection{Implementation details}
For virtual exploration, we sample $N_{\mathrm{Tr}}=20$ virtual trajectories and restrict candidate viewpoints to free space using a rasterization-based occupancy grid with resolution $S=64$.
At each expansion step, we select the top-$3$ candidate views with the highest information gain.
We discretize viewing directions into $D=32$ bins to compute direction-aware coverage gains.

For diffusion-based pseudo-observation generation, we adopt DynamiCrafter~\cite{xing2024dynamicrafter} as the video diffusion backbone and use GPT-4o to generate the text captions for conditioning.
In joint fine tuning stage, we jointly fine-tune the 3DGS model for $15$K iterations using the rendered images, and apply densification during the first $9$K iterations.

\begin{table}[t]
\centering
\vspace{7mm}
\caption{View synthesis results on C3VDv2 under two challenging settings. G.P. denotes GaussianPancakes.}
\label{tab:quantitative_results}
\scriptsize
\setlength{\tabcolsep}{5pt}
\begin{tabular}{l|ccc|ccc}
\hline
\multirow{2}{*}{Method} & \multicolumn{3}{c|}{Sparse: Cecum t1} & \multicolumn{3}{c}{Distanced: Ascending t2} \\
 & PSNR $\uparrow$ & SSIM $\uparrow$ & LPIPS $\downarrow$ & PSNR $\uparrow$ & SSIM $\uparrow$ & LPIPS $\downarrow$ \\
\hline
3DGS & 18.74 & 0.673 & 0.457 & 18.66 & 0.816 & 0.431 \\
G.P. & 18.48 & 0.699 & 0.503 & 10.39 & 0.575 & 0.625 \\
UCNeRF & 19.22 & 0.541 & 0.244 & 20.15 & 0.797 & 0.214 \\
\textbf{Ours} & \textbf{20.23} & \textbf{0.769} & \textbf{0.483} & \textbf{18.75} & \textbf{0.857} & \textbf{0.443} \\
\hline
\end{tabular}
\end{table}

\subsection{Results}
\label{sec:exp_main_results}

\subsubsection{Quantitative Comparison}
\label{sec:nexp_vs_results}
In Table \ref{tab:quantitative_results}, we present the quantitative evaluation of our proposed method, ExtraGS, against state-of-the-art (SOTA) baseline methods, including 3DGS, GaussianPancakes, and UCNeRF. The view synthesis performance is rigorously assessed on the C3VD dataset across two distinctly challenging settings: Scenario 1 (Sparse cecum t1) and Scenario 2 (Distanced ascending t2). To measure the rendering quality and perceptual fidelity, we employ three standard metrics: Peak Signal-to-Noise Ratio (PSNR), Structural Similarity Index (SSIM), and Learned Perceptual Image Patch Similarity (LPIPS).

In Scenario 1, which is characterized by sparse view availability, ExtraGS demonstrates a remarkable advantage in recovering high-fidelity details. Our method achieves the highest PSNR of 20.23 dB and an SSIM of 0.7690, substantially outperforming the closest radiance field baseline, UCNeRF (PSNR 19.22 dB, SSIM 0.5406), especially in preserving structural integrity. Standard Gaussian-based methods such as 3DGS and GaussianPancakes struggle under this sparse setting, yielding lower PSNRs (18.74 dB and 18.48 dB, respectively). This highlights that ExtraGS is markedly more robust when optimizing and rendering scenes from limited observational data.

\newpage
Under Scenario 2, which tests the methods on distanced and complex trajectories, ExtraGS continues to deliver superior performance. It successfully attains the highest SSIM of 0.8566, easily surpassing standard 3DGS (0.8161) and UCNeRF (0.7972). Notably, GaussianPancakes suffers a severe performance degradation in this scenario, with its PSNR dropping to 10.39 dB, illustrating a fundamental inability to generalize across varied viewing distances. Conversely, ExtraGS maintains a highly robust PSNR of 18.75 dB and competitive perceptual quality, validating its stable and consistent view synthesis capabilities regardless of the camera's spatial distribution. 

In conclusion, the quantitative results clearly validate our claims. By securing the best structural similarity (SSIM) and top-tier peak signal-to-noise ratios (PSNR) across both sparse and distanced scenarios, our proposed ExtraGS comprehensively outperforms the SOTA baseline methods. The consistent gains over both standard GS frameworks and modern NeRF variants prove that ExtraGS is a highly effective, versatile, and robust solution for complex novel view synthesis.



\subsubsection{Qualitative Comparison}
\label{sec:exp_exploration_results}
Figure \ref{fig:qualitative_nerfbusters} presents a qualitative evaluation of our proposed method, ExtraGS, against state-of-the-art baseline methods (3DGS, UCNeRF, and GaussianPancakes) on the curated C3VD dataset. The ground truth (GT) endoscopic images reveal highly challenging rendering conditions characterized by complex tubular geometries, intricate tissue folds, fine vascular networks, and dynamic specular highlights.

A visual inspection of the baselines reveals significant limitations in their ability to accurately synthesize these complex biological scenes. Standard 3DGS struggles to maintain smooth textural fidelity, introducing noticeable noisy and grainy artifacts, which are particularly evident in the tissue walls of the bottom row. UCNeRF, while attempting to capture the overall structure, suffers from severe over-smoothing and localized geometric warping (visible in the top row), leading to a washed-out appearance that loses critical high-frequency details. GaussianPancakes demonstrates the most severe instability in this domain; it fails catastrophically in the first row, producing a massive blurring artifact that completely obscures the lumen, and yields overly smoothed, low-fidelity renderings in the subsequent rows.

In contrast, our proposed ExtraGS consistently produces high-fidelity novel views that closely approximate the ground truth. ExtraGS successfully captures the deep topological structures of the tissue folds without the warping or severe blurring exhibited by UCNeRF and GaussianPancakes. Furthermore, our method excels at preserving intricate textural details, accurately reconstructing the fine blood vessels and realistic specular reflections while effectively suppressing the noisy artifacts that plague the standard 3DGS representations.

Overall, by preserving global geometry and fine texture, ExtraGS outperforms the SOTA baselines, proving its robustness and superiority for novel view synthesis in complex environments.


\newpage
\begin{table}[!h]
\centering
\caption{Ablation study on Dataset C3VDv2 cecum t1.}
\label{tab:ablation}
\scriptsize
\setlength{\tabcolsep}{6pt}
\begin{tabular}{@{}l|ccc}
\hline
Variant & PSNR $\uparrow$ & SSIM $\uparrow$ & LPIPS $\downarrow$ \\
\hline
ExtraGS & \textbf{20.38} & \textbf{0.770} & \textbf{0.482} \\
w/o Illum \& Fold aware  & 20.23 & 0.769  & 0.483  \\
\hline
\end{tabular}
\end{table}

\subsection{Ablation Studies}
\label{sec:exp_ablation}

To evaluate the contribution of each design component, we conduct ablation experiments on the \emph{C3VD cecum T1} scene. 
We take the full model \emph{ExtraGS} as the reference and compare it with variants where specific modules are removed while keeping all other settings unchanged, including the data split, initialization, optimization schedule, and evaluation protocol. 
In Table~\ref{tab:ablation}, we mainly investigate the effect of the \emph{illumination- and fold-aware} design by removing this component from the full pipeline. 
Quantitative results are reported using PSNR, SSIM, and LPIPS. 
The full model consistently achieves the best performance (\textbf{20.38} PSNR, \textbf{0.770} SSIM, and \textbf{0.482} LPIPS), outperforming the variant without illumination \& fold awareness (20.23 / 0.769 / 0.483). 
These results suggest that explicitly modeling illumination variation and fold-aware structural cues is beneficial for improving reconstruction quality in challenging endoscopic scenes.


\section{Conclusion}
In this paper, we introduced ExtraGS, a framework designed to address the severe degradation that occurs in endoscopic novel view synthesis under large-baseline camera extrapolation. Instead of relying solely on sparse physical observations, our method integrates 3D Gaussian Splatting with generative diffusion priors to provide additional supervision in poorly observed regions. By identifying anatomical blind spots through uncertainty estimation and refining them via confidence-aware optimization, ExtraGS is able to recover plausible and structurally consistent geometry in areas lacking direct observations. This capability effectively expands the usable field of view and improves reconstruction reliability, suggesting potential benefits for spatial awareness in robot-assisted minimally invasive surgery and preoperative assessment.

Despite these promising results, the current framework assumes relatively static scenes, whereas real surgery involves tissue deformation and dynamic interactions. Our evaluation on offline public datasets does not yet cover dynamic lighting, instrument occlusion, pose estimation errors, unseen tissue types, or real-time execution. Thus, the present results demonstrate potential rather than validated intraoperative reliability. Addressing these limitations, improving computational efficiency, and explicitly modeling tissue deformation will be important for robust 4D reconstruction and future clinical applications.

\clearpage
\bibliographystyle{IEEEtran}
\bibliography{IEEEexample.bib}
\end{document}